\pdfoutput=1

\documentclass[11pt]{article}

\usepackage{emnlp2021}
\usepackage{times}
\usepackage{latexsym}
\usepackage{wrapfig}
\usepackage[T1]{fontenc}

\usepackage[utf8]{inputenc}

\usepackage{microtype}

\usepackage{hyperref}
\usepackage{url}

\usepackage[utf8]{inputenc} 
\usepackage[T1]{fontenc}    
\usepackage{booktabs}       
\usepackage{amsfonts}       
\usepackage{nicefrac}       
\usepackage{microtype}      
\usepackage{graphicx}
\usepackage{amsmath}
\usepackage[font={small}]{caption}

\makeatletter
\def\blfootnote{\xdef\@thefnmark{}\@footnotetext}
\makeatother

\newcommand{\boldx}{\mathbf{x}}
\newcommand{\boldP}{\mathbf{P}}
\newcommand{\boldA}{\mathbf{A}}
\newcommand{\boldV}{\mathbf{V}}
\newcommand{\boldQ}{\mathbf{Q}}
\newcommand{\boldK}{\mathbf{K}}
\newcommand{\boldX}{\mathbf{X}}
\newcommand{\boldH}{\mathbf{H}}

\newcommand{\reals}{\mathbb{R}}

\DeclareMathOperator*{\softmax}{softmax}

\newcommand{\bDiamond}{\mathbin{\Diamond}}

\title{Syntactic Perturbations Reveal Representational Correlates of Hierarchical Phrase Structure in Pretrained Language Models}

  \author{Matteo Alleman$^\dag$ \hspace{3mm} Jonathan Mamou$^{\ddagger}$ \hspace{3mm}  Miguel A Del Rio$^{\bDiamond}$ \hspace{3mm} \\ \bf{Hanlin Tang}$^{\ddagger}$ \hspace{3mm}  \bf{Yoon Kim}$^{\star,\bDiamond,\bullet }$ \hspace{3mm}  \bf{SueYeon Chung}$^{\dag,\bDiamond, \bullet}$ \\
   \\
   $^\dag$Columbia University ~
$^{\ddagger}$Intel  Labs  \\
$^{\star}$MIT-IBM Watson AI ~
$^{\bDiamond}$Massachusetts Institute of Technology }

\begin{document}
\maketitle
\begin{abstract}
\vspace{-5mm}
While vector-based language representations from  pretrained language models have set a new standard for many NLP tasks, there is not yet a complete accounting of their inner workings. In particular, it is not entirely clear what aspects of sentence-level syntax are captured by these representations, nor how (if at all) they are built along the stacked layers of the  network. In this paper, we aim to address such questions with a general class of interventional, input perturbation-based analyses of representations from pretrained language models. Importing from computational and cognitive neuroscience the notion of representational invariance, we perform a series of probes designed to test the sensitivity of these representations to several kinds of structure in sentences. Each probe involves swapping words in a sentence and comparing the representations from perturbed sentences against the original. We experiment with three different perturbations: (1) random permutations of $n$-grams of varying width, to test the scale at which a representation is sensitive to word position; (2) swapping of two spans which do or do not form a syntactic phrase, to test sensitivity to global phrase structure; and (3) swapping of two adjacent words which do or do not break apart a syntactic phrase, to test sensitivity to local phrase structure. 
Results from these probes collectively suggest that Transformers build sensitivity to larger parts of the sentence along their layers, and that hierarchical phrase structure plays a role in this process. 
More broadly, our results also indicate that structured input perturbations widens the scope of analyses that can be performed on often-opaque deep learning systems, and can serve as a complement to existing tools (such as supervised linear probes) for interpreting complex black-box models.\footnote{Datasets, extracted features and code will be publicly available upon publication.} 
{\let\thefootnote\relax\footnote{{$\bullet$ Correspondence}}}
\end{abstract}

\section{Introduction}
\vspace{-3mm}
It is still unknown how distributed information processing systems encode and exploit complex relational structures in data, despite their ubiquitous use in the modern world. The fields of deep learning~\citep{saxe2013learning,stanford_nlp}, neuroscience~\citep{sarafyazd2019hierarchical, stachenfeld2017hippocampus}, and cognitive science~\citep{elman1991distributed, kemp2008discovery, tervo2016toward} have given great attention to this question, including a productive focus on the potential models and their implementations of hierarchical tasks, such as predictive maps and graphs. In this work, we provide a generic means of identifying input structures that deep language models use to ``chunk up'' vastly complex data.


Natural (human) language provides a rich domain for studying how complex hierarchical structures are encoded in information processing systems. More so than other domains, human language is unique in that its underlying hierarchy has been extensively studied and theorized in linguistics, which provides source of ``ground truth'' structures for stimulus data. Much prior work on characterizing the types of linguistic information encoded in computational models of language such as neural networks has focused on supervised readout probes, which train a classifier on top pretrained models to predict a particular linguistic label \citep{belinkov2017analyzing,liu_et_al,tenney2019bert}. In particular, \citet{stanford_nlp} apply probes to discover linear subspaces that encode tree-distances as distances in the representational subspace, and \citet{kim2019pre} show that these distances can be used even without any labeled information to induce hierarchical structure. However, recent work has highlighted issues with correlating supervised probe performance with the amount of language structure encoded in such representations \citep{hewitt2019}. Another popular approach to analyzing deep models is through the lens of geometry \citep{reif2019visualizing,gigante2019visualizing}. While geometric interpretations provide significant insights, they present another challenge in  summarizing the structure in a quantifiable way. More recent techniques such as replica-based mean field manifold analysis method \citep{chung2018classification, cohen2019separability, mamou2020emergence} connects representation geometry with linear classification performance, but the method is limited to categorization tasks. 

\begin{figure}[t!]
    \centering
     \includegraphics[width=\linewidth]{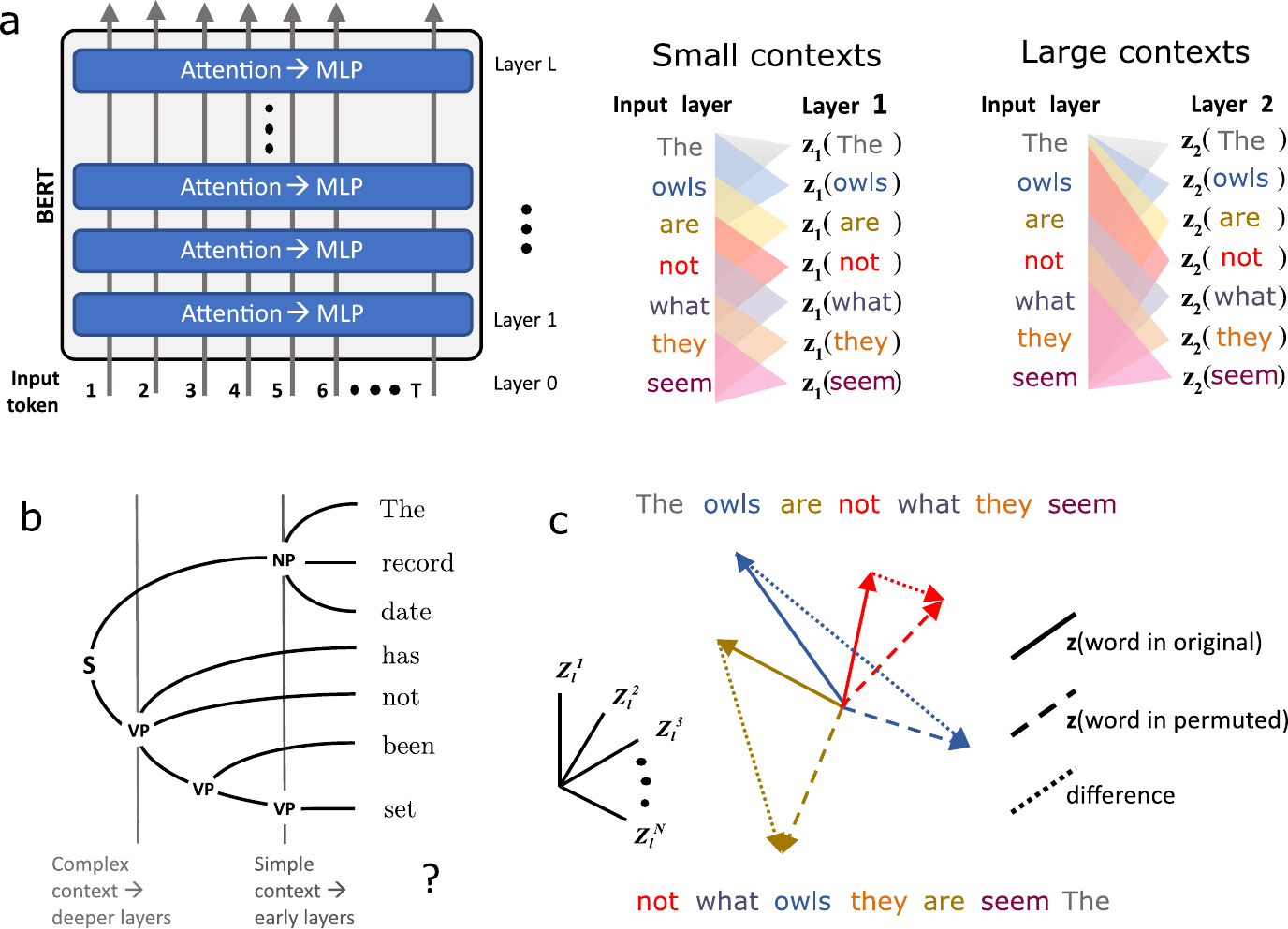}
     \vspace{-4mm}
    \caption{Do Transformers build complexity along their layers? (\textbf{a}) The representation of a word is a function of its context, and this cartoon illustrates an hypothesis that deeper representations use larger contexts. (\textbf{b}) An example parse tree, illustrating our notion of phrase complexity. (\textbf{c}) Cartoon of the distortion metric, where vectors are the z-scored feature vectors $\mathbf{z}$, and color map vectors to words.}
    \label{fig:fig1}
    \vspace{-5mm}
\end{figure}

In this work, we make use of an experimental framework from cognitive science and neuroscience to probe for hierarchical structure in contextual representations from pretrained Transformer models (i.e., BERT~\citep{bert} and its variants). A popular technique in neuroscience involves measuring change in the population activity in response to controlled, input perturbations~\citep{mollica2020composition, ding2016cortical}.  We apply this approach to test the characteristic scale and the complexity (Fig.~\ref{fig:fig1}) of hierarchical phrase structure encoded deep contextual representations, and present several key findings:



\begin{enumerate}
    \item Representations are distorted by shuffling small $n$-grams in early layers, while the distortion caused by shuffling large $n$-grams starts to occur in later layers, implying the scale of characteristic word length increases from input to downstream layers. 
    \item Representational distortion caused by swapping two constituent phrases is smaller than when the control sequences of the same length are swapped, indicating that the BERT representations are sensitive to hierarchical phrase structure.
    \item Representational distortion caused by swapping adjacent words across phrasal boundary is larger than when the swap is within a phrasal boundary; furthermore, the amount of distortion increases with the syntactic distance between the swapped words. The correlation between distortion and tree distance increases across the layers, suggesting that the characteristic complexity of phrasal subtrees increases across the layers.
    \item Early layers pay more attention between syntactically closer adjacent pairs and deeper layers pay more attention between syntactically distant adjacent pairs. The attention paid in each layer can explain some of the emergent sensitivity to phrasal structure across layers.
\end{enumerate}


Our work demonstrates that interventional tools such as controlled input perturbations can be useful for analyzing deep networks, adding to the growing, interdisciplinary body of work which profitably adapt  experimental techniques from cognitive neuroscience and psycholinguistics to analyze computational models of language~\citep{futrell2018rnn,wilcox2019hierarchical,futrell-etal-2019-neural,ettinger2020bert}.
\vspace{-1mm}
\section{Methods}
\vspace{-1mm}
\label{methods}
Eliciting changes in behavioral and neural responses through controlled input perturbations  is a common experimental technique in cognitive neuroscience and psycholinguistics~\citep{tsao2008mechanisms,mollica2020composition}. Inspired by these approaches, we perturb input sentences and measure the discrepancy between the resulting, perturbed representation  against the original. While conceptually simple, this approach allows for a targeted analysis of internal representations obtained from different layers of deep models, and can suggest partial mechanisms by which such models are able to encode linguistic structure. We note that sentence perturbations have been primarily utilized in NLP for representation learning~\citep{hill-etal-2016-learning-distributed,Artetxe2018mt,lample2018mt}, data augmentation \citep{wang-etal-2018-switchout,andreas2020geca}, and testing for model robustness (e.g., against adversarial examples)~\citep{jia-liang-2017-adversarial,belinkov2018noise}. A methodological contribution of our work is to show that input perturbations can complement existing tools and widens the scope of questions that could be asked of representations learned by deep networks.

\vspace{-1mm}
\subsection{Sentence perturbations}
\vspace{-1mm}
In this work we consider three different types of sentence perturbations designed to probe for different phenomena.
\vspace{-1mm}
\paragraph{$n$-gram shuffling} In the $n$-gram shuffling experiments, we randomly shuffle the words of a sentence in units of $n$-grams, with $n$ varying from 1 (i.e., individual words) to 7 (see Fig.~\ref{fig:fig2}a for an example). While the number of words which change absolute position is similar for different $n$, larger $n$ will better preserve the local context (i.e., relative position) of more words. Thus, we reason that $n$-gram swaps affect the representations selective to the context with size $n$ or higher within the sentence, and that lower $n$ will result in greater distortion in sentence representations.
\vspace{-1mm}
\paragraph{Phrase swaps} The $n$-gram shuffling experiments probe for sensitivity of representations to local context without taking into account syntactic structure. In the phrase swap experiments, we perturb a sentence by swapping two randomly chosen spans. We explore two ways of swapping spans. In the first setting, the spans are chosen such that they are valid phrases according to its parse tree.\footnote{We use constituency parse trees from the English Penn Treebank \citep{ptb}.} In the second setting, the spans are chosen that they are invalid phrases. Importantly, in the second, control setting, we fix the length of the spans  such that the lengths of spans that are chosen to be swapped are the same as in the first setting (see Fig.~\ref{fig:fig3}a for an example). We hypothesize that swapping invalid phrases will result in more distortion than swapping valid phrases, since invalid swaps will result in greater denigration of syntactic structure. 
\vspace{-1mm}
\paragraph{Adjacent word swaps} In the adjacent word swapping experiments, we swap two adjacent words in a sentence. We again experiment with two settings -- in the first setting, the swapped words stay within the  phrase boundary (i.e., the two words share the same parent), while in the second setting, the swapped words cross phrase boundaries. 
We also perform a more fine-grained analysis where we condition the swaps based on the ``syntactic distance'' between the swapped words, where syntactic distance is defined as the distance between the two words in the parse tree (see Fig.~\ref{fig:fig4}c). Since a phrase corresponds to a subtree of the parse tree, this distance also quantifies the number of nested phrase boundaries between two adjacent words. Here, we expect the amount of distortion to be positively correlated with the syntactic distance of the words that are swapped.
\vspace{-2mm}
\subsection{Contextual representations from Transformers}
\vspace{-1mm}
For our sentence representation, we focus on the Transformer-family of models pretrained on large-scale language datasets (BERT and its variants).
Given an input word embedding matrix $\boldX \in \reals^{T \times d}$ for a sentence of length $T$, the Transformer applies self attention over the  previous layer's representation to produce a new representation, 
\begin{equation}
\begin{aligned}
    &\boldX_l = f_{l}([\boldH_{l,1}, \dots, \boldH_{l, H}]), \hspace{2mm} \boldH_{l, i} = \boldA_{l,i} \boldX_{l-1} \boldV_{l,i}, \\ &\boldA_{l,i} = \softmax\left(\frac{(\boldX_{l-1}\boldQ_{l,i})(\boldX_{l-1}\boldK_{l,i})^\top}{\sqrt{d_k}}\right),
    \label{eq:trans}
\end{aligned}
\end{equation}
where $f_{l}$ is an MLP layer, $H$ is the number of heads, $d_H = \frac{d}{H}$ is the head embedding dimension, and $\boldQ_{l,i}, \boldK_{l,i}, \boldV_{l,i} \in \reals^{d \times d_k}$ are respectively the learned query, key, and value projection matrices at layer $l$ for head $i$. The MLP layer consists of a residual layer followed by layer normalization and a nonlinearity. The $0$-th layer representation $\boldX_0$ is obtained by adding the position  embeddings and the segment embeddings to the input token embeddings $\boldX$, and passing it through normalization layer.\footnote{However, the exact specification for the MLP and $\boldX_0$ may vary across different pretrained models.}

In this paper, we conduct our distortion analysis mainly on the intermediate Transformer representations $\boldX_{l}=[\boldx_{l,1}, \dots, \boldx_{l, T}]$, where $\boldx_{l, t} \in \reals^{d} $ is the contextualized representation for word $t$ at layer $l$.\footnote{BERT uses BPE tokenization~\citep{sennrich2015neural}, which means that some words are split into multiple tokens. Since we wish to evaluate  representations at word-level, if a word is split into multiple tokens, its word representation is computed as the average of all its token representations.} We analyze the trend in distortion as a function of layer depth $l$ for the different perturbations. We also explore the different attention heads $\boldH_{l,i} \in \reals^{T \times d_{H}}$ and the associated attention matrix $\boldA_{l,i} \in \reals^{T \times T}$ to inspect whether certain attention heads specialize at encoding syntactic information.


\vspace{-1mm}
\subsection{Distortion metric}
\vspace{-1mm}
Our input manipulations allow us to specify the distortion at the input level, and we wish to measure the corresponding distortion in the representation space (Fig.~\ref{fig:fig1}). Due to the attention mechanism, a single vector in an intermediate layer is a function of the representations of (potentially all) the other tokens in the sentence. Therefore, the information about a particular word might be distributed among the many feature vectors of a sentence, and we wish to consider all feature vectors together as a single sentence-level representation.



We thus represent each sentence as a matrix and use the distance induced by matrix 2-norm. Specifically, let $\boldP \in \{0,1\}^{T \times T}$ be the binary matrix representation of a permutation that perturbs the input sentence, i.e., $\tilde{\boldX} = \boldP\boldX$. Further let $\tilde{\boldX}_{l}$ and $\boldX_{l}$ be the corresponding sentence representations for the $l$-th layer for the perturbed and original sentences. To ignore uniform shifting and scaling, we also z-score each feature dimension of each layer (by subtracting the mean and dividing by the standard deviation where these statistics are obtained from the full Penn Treebank corpus) to give $\tilde{\mathbf{Z}}_l$ and $\mathbf{Z}_l$.
Our distortion metric for layer $l$ is then defined as $\Vert \mathbf{Z}_l - \boldP^{-1}\tilde{\mathbf{Z}}_l \Vert /\sqrt{Td}$, where $\Vert \cdot \Vert$ is the matrix 2-norm (i.e., Frobenius norm).\footnote{There are many possible ways of measuring distortion, such as the average cosine similarity or distance between corresponding feature vectors, as well as different matrix norms. We observed the results to be qualitatively similar for different measures, and hence we focus on the Frobenius norm in our main results. We show the results from additional distortion metrics in Sec.~\ref{SM:metrics}.} Importantly, we invert the permutation of the perturbed representation  to preserve the original ordering, which allows us to measure the distortion due to structural change, rather than distortion due to simple differences in surface form. We divide by $\sqrt{Td}$ to make the metric comparable between sentences (with different $T$) and networks (with different $d$).

Intuitively, our metric is the scaled Euclidean distance between the z-scored, flattened sentence representations, $\mathbf{z}_l \in \reals^{Td}$. Because each dimension is independently centered and standardized, the maximally unstructured distribution of $\mathbf{z}_l$ is an isotropic $Td$-dimensional Gaussian. The expected distance between two such vectors is approximately $\sqrt{2Td}$. Therefore, we can interpret a distortion value approaching $\sqrt{2}$ as comparable to if we had randomly redrawn the perturbed representation.
\vspace{-1mm}
\subsubsection{Additional metrics}\vspace{-1mm}
\label{SM:metrics}
In addition to the scaled Frobenius distance, we also considered other ways of measuring distortion in the representation. We will briefly report results for two other metrics, and describe them here.
\vspace{-1mm}
\paragraph{CCA} Canonical correlations analysis (CCA)~\cite{svcca} measures the similarity of two sets of variables using many samples from each. Given two sets of random variables $\mathbf{x}=(x_1, x_2, ..., x_n)$ and $\mathbf{y}=(y_1, y_2, ..., y_m)$, CCA finds linear weights $\mathbf{a}\in \reals^n$ and $\mathbf{b} \in \reals^m$ which maximise $\mathrm{cov} (\mathbf{a} \cdot \mathbf{x}, \mathbf{b} \cdot \mathbf{y})$. In our context, we treat the representation of the original sentence as $\mathbf{x}$, and the representation of the perturbed sentence as $\mathbf{y}$, and the resulting correlation as a similarity measure. 

Since CCA requires many samples, we use the set of all word-level representations across all perturbed sentences. For example, to construct the samples of $\mathbf{x}$ from $S$ perturbed sentences, we get use $\left[ \mathbf{X}_1 \vert \mathbf{X}_2 \vert ... \vert \mathbf{X}_S \right]$, where each $\mathbf{X}_i \in \reals^{768 \times T_i}$. Unless specified otherwise, $S=400$. For good estimates, CCA requires many samples (on the order of at least the number of dimensions), and we facilitate this by first reducing the dimension of the matrices using PCA. Using 400 components preserves $\sim90 \%$ of the variance. Thus, while CCA gives a good principled measure of representational similarity, its hunger for samples makes it unsuitable as a per-sentence metric.

We also measured distortion using Projection Weighted Canonical Correlation Analysis (PWCCA), an improved version of CCA to estimate the true correlation between tensors~\cite{pwcca}.\footnote{For both CCA and PWCCA, we use the implementation from \url{https://github.com/google/svcca}.} 

As reported in Figure~\ref{fig:fig5}, we did not find any qualitative differences between PWCCA and CCA in our experiments. 
\vspace{-1mm}
\paragraph{Cosine} A similarity measure defined on individual sentences is the cosine between the sentence-level representations. By sentence-level representation, we mean the concatenation of the word-level vectors into a single vector $\mathbf{s} \in \reals^{NT}$ (where $N$ is the dimension of each feature vector). Treating each dimension of the vector as a sample, we can then define the following metric: $\mathrm{corr}\left( \mathbf{s}^{original}_i, \mathbf{s}^{swapped}_i \right)$. This is equivalent to computing the cosine of the vectors after subtracting the (scalar) mean across dimensions, hence we will refer to it as `cosine'.
\vspace{-1mm}
\subsection{Model details}
\vspace{-1mm}
\label{SM:models}
Here we give the details for all models considered in this paper. The majority of results are from BERT, but we also tested other variants.\footnote{We use the implementation from \url{https://github.com/huggingface/transformers}.}
\begin{itemize}
    \item {\bf BERT} \cite{bert} {\tt bert-base-cased}. 12-layer, 768-hidden, 12-heads, 110M parameters.
    \item {\bf RoBERTa} \cite{liu2019roberta} {\tt roberta-base}. 12-layer, 768-hidden, 12-heads, 125M parameters.
    \item {\bf ALBERT} \cite{lan2019albert} {\tt albert-base-v1}. 12 repeating layers, 128 embedding, 768-hidden, 12-heads, 11M parameters.
    \item {\bf DistilBERT} \cite{sanh2019distilbert} {\tt distilbert-uncased}. 6-layer, 768-hidden, 12-heads, 66M parameters.
The model distilled from the BERT model {\tt bert-base-uncased} checkpoint.
    \item {\bf XLNet} \cite{yang2019xlnet} {\tt xlnet-base-cased}. 12-layer, 768-hidden, 12-heads, 110M parameters.
\end{itemize}
Note that the hidden size is 768 across all the models.
For each pre-trained model, input text is tokenized using its default tokenizer and features are extracted at token level. 
\vspace{-1mm}
\section{Experimental Setup}
\vspace{-1mm}
We apply our perturbation-based analysis on sentences from the English Penn Treebank~\citep{ptb}, where we average the distortion metric across randomly chosen sentences (see Sec.~\ref{SM:dataset} for the exact details). We analyze the distortion, as measured by length-normalized Frobenius norm between the perturbed and original representations, as a function of layer depth. Layers that experience large distortion when the syntactic structure is disrupted from the perturbation can  be interpreted as being more sensitive to hierarchical syntactic structure. 

As we found the trend to be largely similar across the different models, in the following section, we primarily discuss results from BERT ({\tt bert-base-cased}). We replicate key results with other pretrained and randomly-initialized Transformer-based  models as well.

\section{Results}
\label{results}

We summarize our findings for the different perturbations below. While not shown in the main results, we note that  randomly-initialized (i.e. untrained) models (somewhat unsuprisingly) exhibit a flat distortion trend for all perturbations (see Sec.~\ref{SM:models}). This indicates that the patterns observed here are due to the model's structural knowledge acquired through training, and not simply due to the underlying architecture. 
\subsection{Characteristic scale increases along BERT layers}

\begin{figure}[ht]
    \centering
    \includegraphics[width=\linewidth]{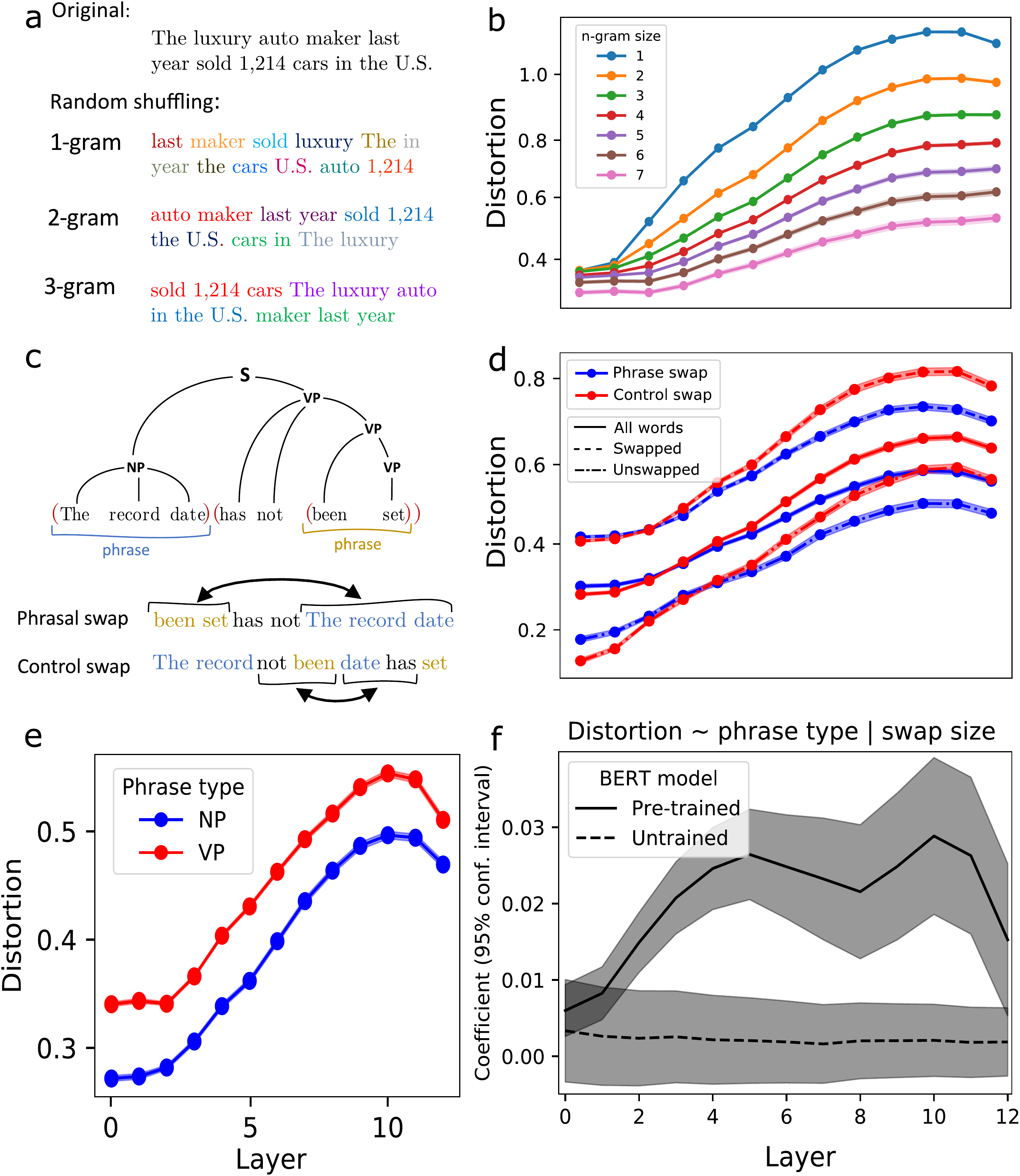}
    \vspace{-6mm}
    \caption{Swapping $n$-grams and phrases. (\textbf{a}) Examples of basic $n$-gram shuffles, where colors indicate the units of shuffling. (\textbf{b}) Distortion metric computed at each layer, conditioned on $n$-gram size. Error bars hereafter represent standard error across 400 examples. (\textbf{c}) An example parse tree, with phrase boundaries shown as grey brackets, and two low-order phrases marked; and examples of a phrasal and control swap, with colors corresponding to the phrases marked above. (\textbf{d}) Distortion, computed at each layer, using either the full sentence, the subsentence of unswapped words, or the subsentence of swapped words, conditioned on swap type. (e) Full-sentence distortion for VP and NP phrase swaps. (f) Partial linear regression coefficients (see \ref{SM:pmi}) for pre-trained and untrained BERT models after controlling for swap size.}
    \label{fig:fig2}
    \vspace{-5mm}
\end{figure}

When we shuffle in units of larger $n$-grams, it only introduces distortions in the deeper BERT layers compared to smaller $n$-gram shuffles. The $n$-gram sized shuffles break contexts larger than $n$, while preserving contexts of size $n$ or smaller. Interestingly, smaller $n$-gram shuffles diverge from the original sentence in the early layers (Fig.~\ref{fig:fig2}b, top curve), implying that only in early layers are representations built from short-range contexts. Larger $n$-gram shuffles remain minimally distorted for `longer' (Fig.~\ref{fig:fig2}b, bottom curve), implying that long-range contexts play a larger role deeper layer representations.
\paragraph{Phrasal boundaries matter}
Since BERT seems to build larger contexts along its layers, we now ask whether those contexts are structures of some grammatical significance. A basic and important syntactic feature is the constituent phrase, which BERT has previously been shown to represented in some fashion~\citep{goldberg2019assessing,kim2019pre}. We applied two targeted probes of phrase structure in the BERT representation, and found that phrasal boundaries are indeed influential.

\begin{figure}[ht]
    \centering
    \includegraphics[width=\linewidth]{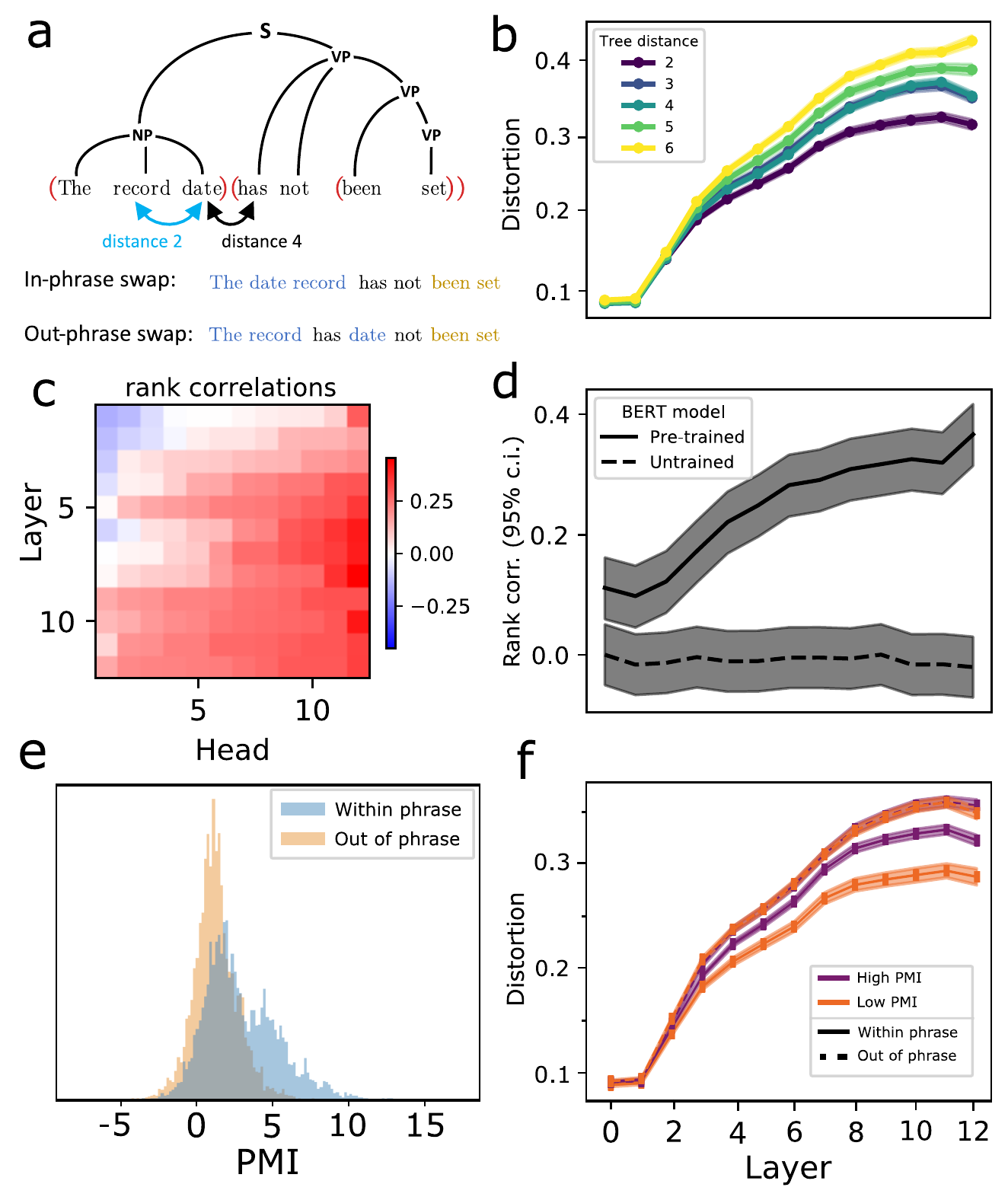}
    \vspace{-5mm}
    \caption{Syntactic distance affects representational distortion. (\textbf{a}) An example of adjacent swaps which do and do not cross a phrase boundary, with low-order phrases colored. Phrase boundaries are drawn in red. (\textbf{b}) Distortion in each layer, but conditioned on the tree distance. (\textbf{c}) For each head (column) of each layer (row), the (Spearman) rank correlation between distortion and tree distance of the swapped words. Colors are such that red is positive, blue negative. (\textbf{d}) Rank correlations between distortion (of the full representation) in the trained and untrained BERT models. (\textbf{e}) Histogram of PMI values, for pairs in the same phrase and not. (\textbf{f}) Similar to \textbf{b}, but averaging all out-of-phrase swaps, and separating pairs above (`high') or below (`low') the median PMI.}
    \vspace{-5mm}
    \label{fig:fig3}
\end{figure}

If we swap just two $n$-grams, the BERT representations are less affected when phrases are kept intact. We show this by swapping only two $n$-grams per sentence and comparing the distortion when those $n$-grams are phrases to when they cross phrase boundaries (Fig.~\ref{fig:fig3}a), where we control for the length of $n$-grams that are swapped in both settings. There is less distortion when respecting phrase boundaries, which is evident among all feature vectors, including those in the position of words which did not get swapped (Fig.~\ref{fig:fig2}d). The global contextual information, distributed across the sentence, is affected by the phrase boundary.

Swapping verb phrases (VP) also results in more distortion than swapping noun phrases (NP) (Fig.~\ref{fig:fig2}e). Since VP are in general larger than NP, this effect could in principle be due simply to the number of words being swapped. Yet that is not the case: Using a partial linear regression (see details in \ref{SM:pmi}), we can estimate the difference between the VP and NP distortions conditional on any smooth function of the swap size, and doing this reveals that there is still a strong difference in the intermediate layers (Fig.~\ref{fig:fig2}f).

\subsection{Phrase hierarchy matters}
Having seen that representations are sensitive to phrase boundaries, we next explore whether that sensitivity is proportional to the number of phrase boundaries that are broken, which is a quantity related to the phrase hierarchy. Instead of swapping entire phrases, we swap two adjacent words and analyze the distortion based on how far apart the two words are in the constituency tree (Fig.~\ref{fig:fig3}a)\footnote{Note that for adjacent words, the number of broken phrase boundaries equals the tree distance minus two.}. This analysis varies the distance in the deeper tree structure while keeping the distance in surface form constant (since we always swap adjacent words).

If the hierarchical representations are indeed being gradually built up along the layers of these pretrained models, we expect distortion to be greater for word swaps that are further apart in tree distance. We indeed find that there is a larger distortion when swapping syntactically distant words (Fig.~\ref{fig:fig3}b). This distortion  grows from earlier to later BERT layers. Furthermore, when looking at the per-head representations of each layer, we see that in deeper layers there are more heads showing a positive rank correlation between distortion and tree distance (Fig.~\ref{fig:fig3}c). In addition to a sensitivity to phrase boundaries, deeper BERT layers develop a sensitivity to the number of boundaries that are broken.

\begin{figure}
    \centering
    \includegraphics[width=\linewidth]{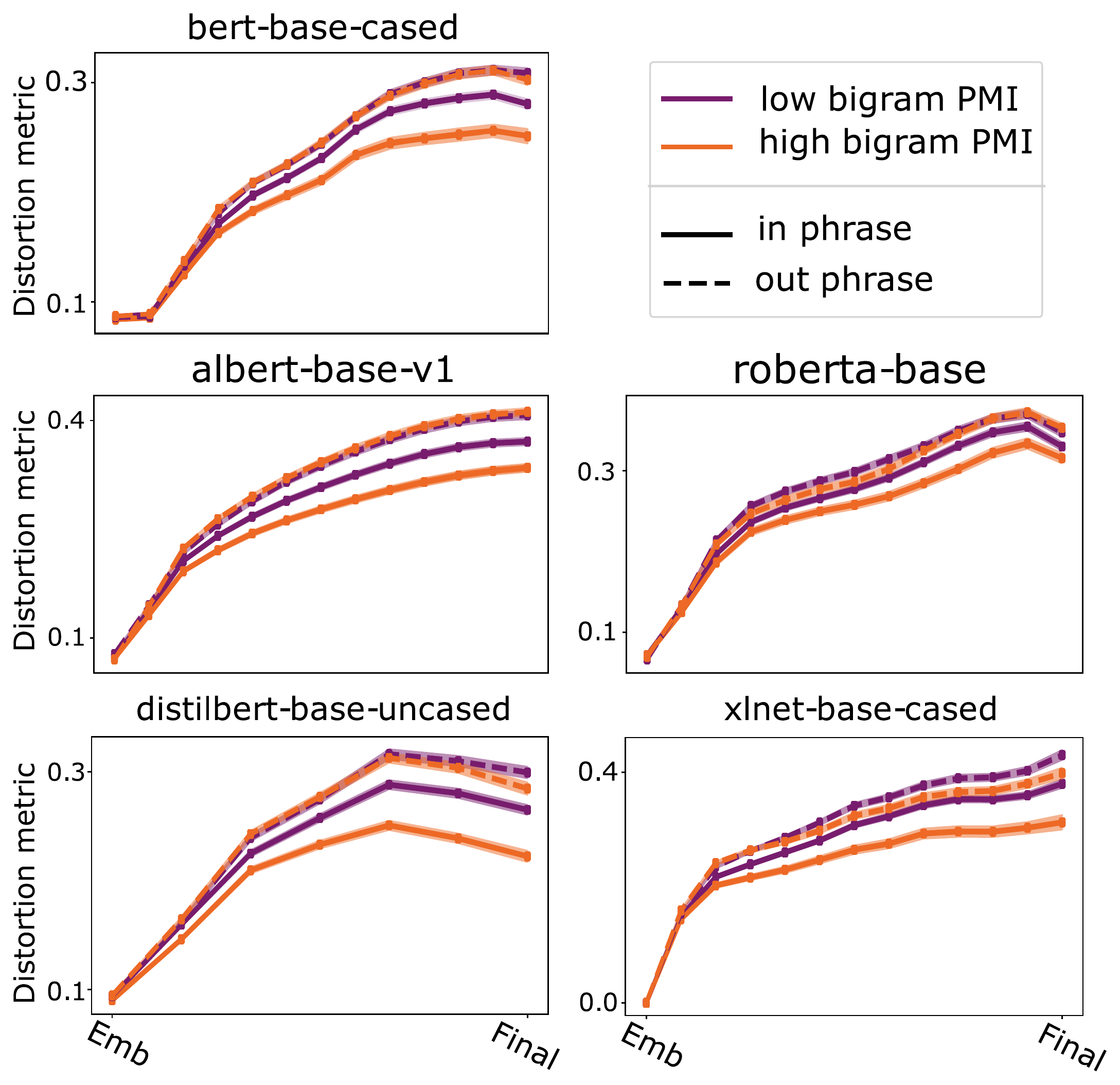}
    \vspace{-5mm}
    \caption{Replicating the adjacent word swapping experiments using different transformer architectures. Lines are the mean Frobenius distance, and the shading is $\pm$1 standard error of the mean.}
    \label{fig:fig4}
    \vspace{-5mm}
\end{figure}

\paragraph{Controlling for co-occurrence}
Since words in the same phrase may tend to occur together more often, co-occurrence is a potential confound when assessing the effects of adjacent word swaps. Co-occurrence is a simple statistic which does not require any notion of grammar to compute -- indeed it is used to train many non-contextual word embeddings (e.g., word2vec~\cite{mikolov2013distributed}, GloVe~\cite{pennington2014glove}). So it is natural to ask whether BERT's resilience to syntactically closer swaps goes beyond simple co-occurrence statistics. For simplicity, let us focus on whether a swap occurs within a phrase (tree distance = 2) or not.

As an estimate of co-occurrence, we used the pointwise mutual information (PMI). Specifically, for two words $w$ and $v$, the PMI is
$\log \frac{p(w,v)}{p(w) p(v)}$, which is
estimated from the empirical probabilities. We confirm that adjacent words in the same phrase do indeed have a second mode at high PMI (Fig.~\ref{fig:fig3}e). Dividing the swaps into those whose words have high PMI (above the marginal median) and low PMI (below it), we can see visually that the difference between within-phrase swaps and out-of-phrase swaps persists in both groups (Fig.~\ref{fig:fig3}f). For a more careful statistical test, in the appendix we show results from running a linear regression between distortion and the phrase boundary which accounts for dependency on any smooth function of PMI (the same method as used in the VP/NP swaps; see details in \ref{SM:pmi}). Even when accounting for the effect of PMI, there is a significant correlation between the breaking of a phrase and the subsequent distortion. This indicates that the greater distortion for word swaps which cross phrase boundaries is not simply due to surface co-occurrence statistics.
\begin{figure}
    \centering
    \includegraphics[width=\linewidth]{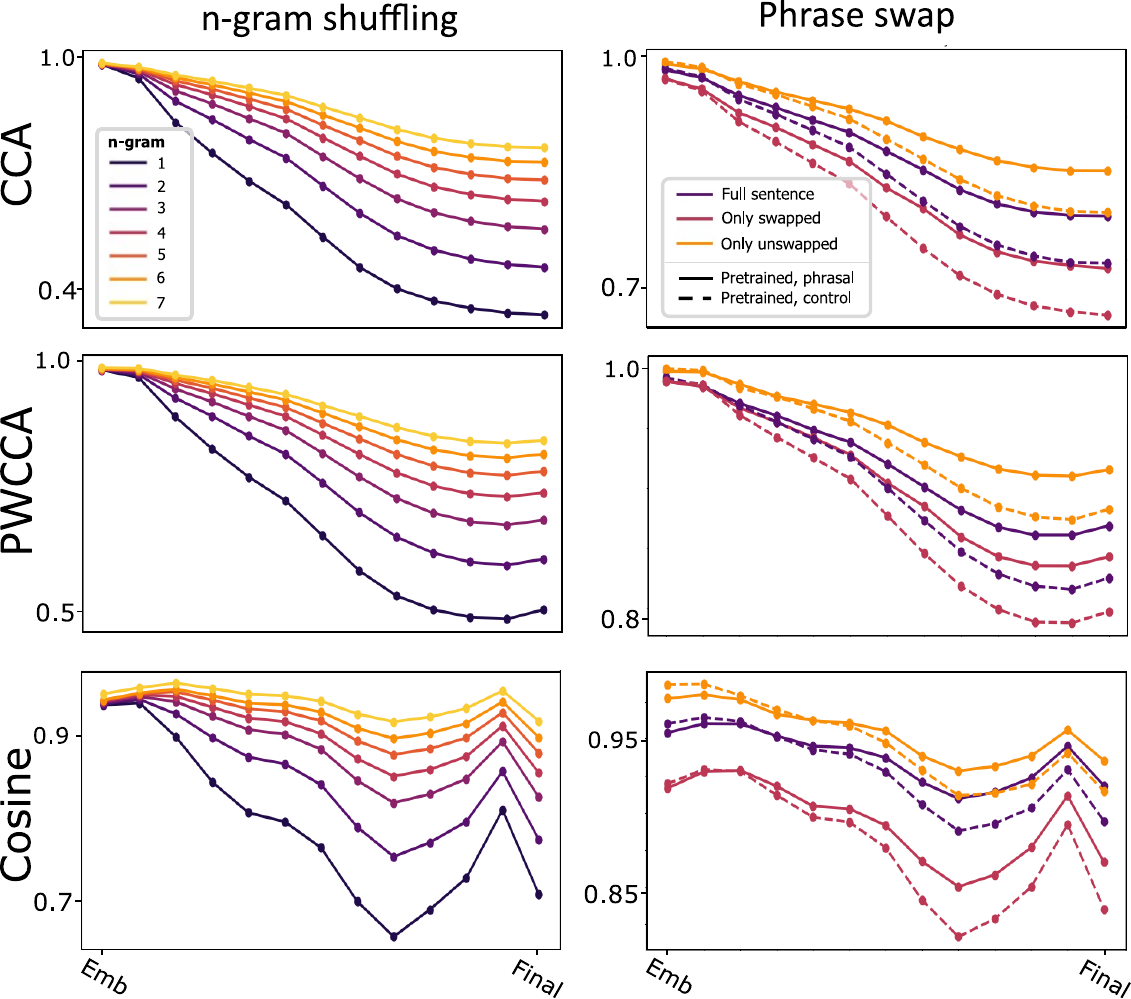}
    \vspace{-5mm}
    \caption{Results from the pretrained BERT model using alternative distortion metrics, on the $n$-gram shuffling and phrase swap experiments.}
    \label{fig:fig5}
\vspace{-5mm}
\end{figure}
\paragraph{Other models and metrics}
To confirm that these basic effects are not idiosyncrasies of the BERT model and our chosen distortion metric, we replicated the previous results in other transformer models and with different metrics (see \ref{SM:metrics} for details). For economy of space, we report only the adjacent word swaps in alternative models (Fig.~\ref{fig:fig4}) and the n-gram and phrase swap experiments with alternative metrics (Fig.~\ref{fig:fig5}). In all tested models, using the initialization weights (which preserve the architecture, but are untrained) resulted in distortions which were constant across layers and perturbation type, indicating that the observed sensitivities cannot be explained by transformer architecture alone.

\begin{figure*}[t]
    \centering
    \includegraphics[width=0.9\linewidth]{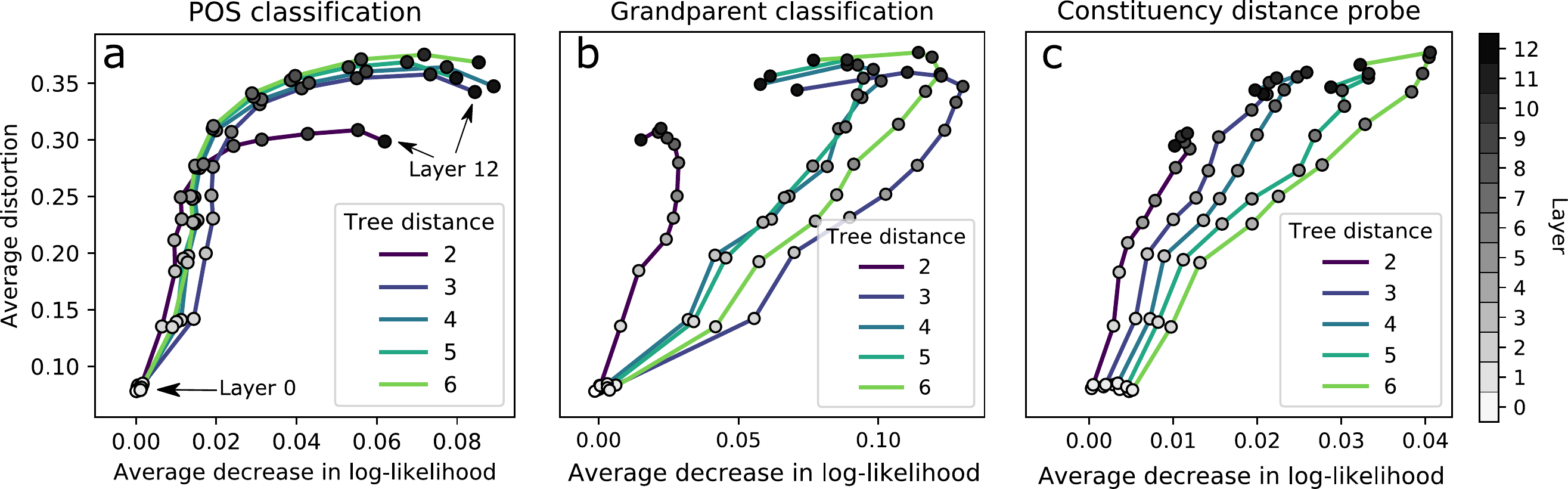}
\vspace{-2mm}
    \caption{Distortion and inference impairment for increasing linguistic complexity. In each plot, a point is the average (distortion, `impairment') for a given layer and a given class of word swap distance. Points are connected by lines according to their swap type (i.e. tree distance). The circles are colored according to layer (see right for a legend). Averages are taken over 600 test sentences, with one of each swap type per sentence, and both distortion and log-likelihood are computed for every word in the sentence.}
    \label{fig:fig6}
    \vspace{-5mm}
\end{figure*} 

\paragraph{Effects on linguistic information}
Do our input perturbations, and the resulting the distortions, reflect changes in the encoding of important linguistic information? One way to address this question, which is popular in computational neuroscience \citep{dicarlo2007untangling} and more recently BERTology \citep{liu_et_al,tenney2019bert}, is to see how well a linear classifier trained on a linguistic task generalizes from the (representations of the) unperturbed sentences to the perturbed ones. With supervised probes, we can see how much the representations change with respect to the subspaces that encode specific linguistic information.

Specifically, we relate representational distortion to three common linguistic tasks of increasing complexity: part of speech (POS) classification; grandparent tag (GP) classification \citep{tenney2019bert}; and a parse tree distance reconstruction \citep{stanford_nlp}\footnote{While the original paper predicted dependency tree distance, in this paper we instead predict the constituency tree distance.}. The probe trained on each of these tasks is a generalized linear model, linearly mapping a datapoint $\textbf{x}$ (i.e. BERT representations from different layers) to a conditional distribution of the labels, $p(y \vert \theta(\textbf{x}))$ (see \ref{SM:probes} for more details). Thus a ready measure of the effect of each type of swap, for a single sentence, is $\log p(y \vert \theta(\textbf{x}_i)) - \log p(y \vert \theta(\tilde{\textbf{x}}_i))$, where $\tilde{\textbf{x}}_i$ is same datum as $\textbf{x}_i$ in the perturbed representation\footnote{POS- and GP-tag prediction outputs a sequence of labels for each sentence, while the distance probe outputs the constituency tree distance between each pair of words. Then $\log p(y|\theta(\mathbf{x}_i))$ is simply the log probability of an individual label.}. Averaging this quantity over all datapoints gives a measure of content-specific distortion within a representation, which we will call ``inference impairment''.

Based on the three linguistic tasks, the distortion we measure from the adjacent word swaps is more strongly related to more complex information. The inverted L shape of Fig. \ref{fig:fig6}a suggests that increasing distortion is only weakly related to impairment of POS inference, which is perhaps unsurprising given that POS tags can be readily predicted from local context. A deeper syntactic probe, the GP classifier (\ref{fig:fig6}b), does show a consistent positive relationship, but only for swaps which break a phrase boundary (i.e. distance >2). Meanwhile, impairment of the distance probe (\ref{fig:fig6}c), which reconstructs the full parse tree, has a consistently positive relationship with distortion, whose slope is proportionate to the tree distance of the swap. Thus, when specifically perturbing the phrasal boundary, the representational distortion is related to relatively more complex linguistic information.

\subsection{Investigation of attention}
In the transformer architecture, contexts are built with the attention mechanism. Recall that attention is a mechanism for allowing input vectors to interact when forming the output, and the ultimate output for a given token is a convex combination of the features of all tokens (Eq.~\ref{eq:trans}). It has been shown qualitatively that, within a layer, BERT allocates attention preferentially to words in the same phrase~\citep{kim2019pre}, so if our perturbations affect inference of phrase structure then the changes in attentions could explain our results. Note that it is not guaranteed to do so: the BERT features in a given layer are a function of the attentions and the ``values'' (each token's feature vector), and both are affected by our perturbations. Therefore our last set of experiments asks whether attention alone can explain the sensitivity to syntactic distance.

To quantify the change in attention weights across the whole sentence, we compute the distance between each token's attention weights in the perturbed and unperturbed sentences, and average across all tokens. For token $i$, its vector of attention weights in response to the unperturbed sentence is $a^i$, and for the perturbed one $\tilde{a}^i$ (such that $\sum_j a^i_j = 1$). Since each set of attention weights are non-negative and sum to 1 due to softmax, we use the relative entropy\footnote{Also called the KL divergence for probability distributions.} as a distance measure. This results in the total change in attention being: \[\Delta a = \frac{1}{T} \sum_{i=1}^{T} \sum_{j=1}^{T} a^i_j \log \frac{a^i_j}{\tilde{a}^i_j}\] which is non-negative and respects the structure of the weights. We confirmed that other measures (like the cosine similarity) produce results that are qualitatively similar.

\begin{figure}
    \centering
    \includegraphics[width=\linewidth]{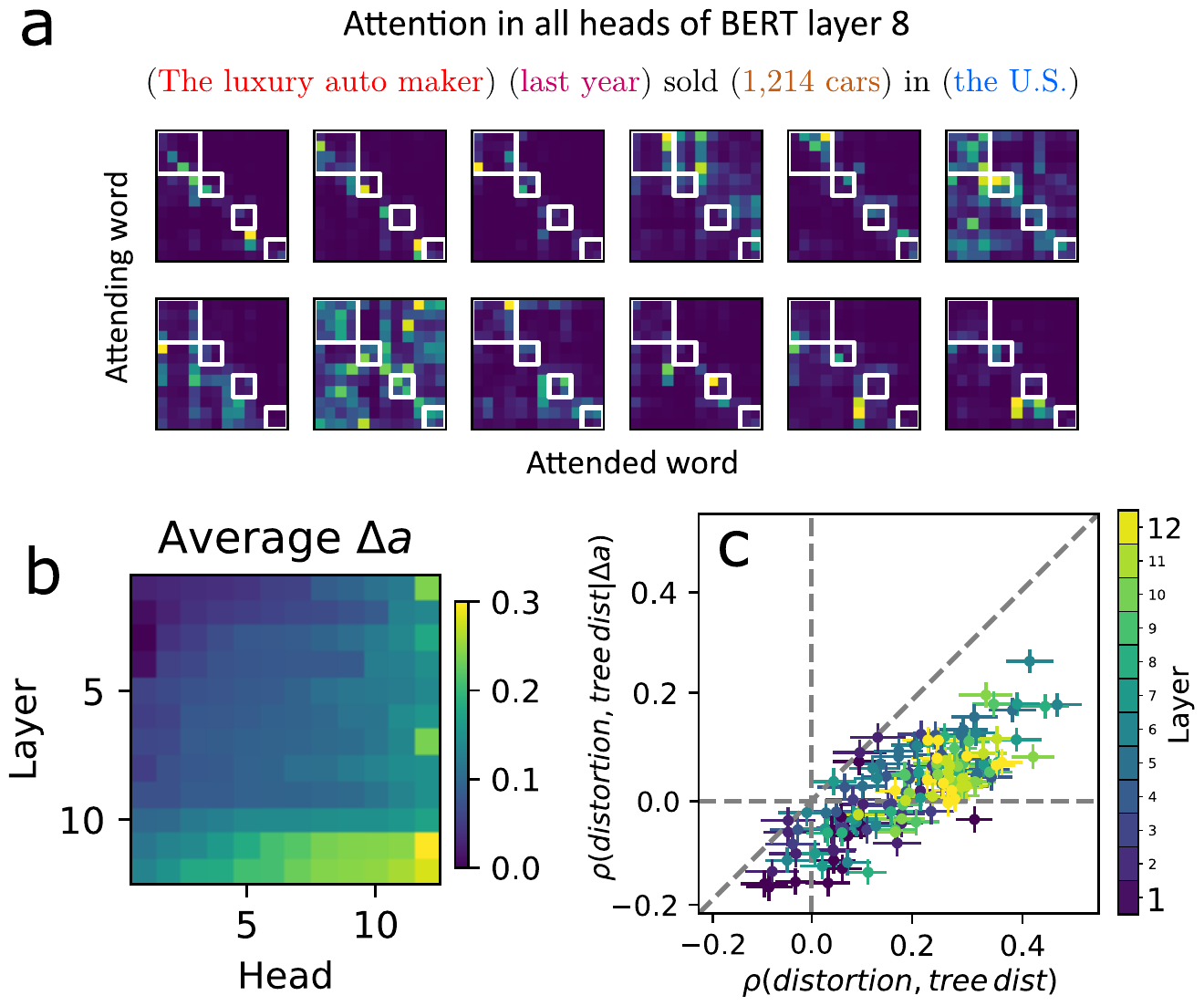}
    \vspace{-5mm}
    \caption{Attention alone explains part of the sensitivity to tree distance. (\textbf{a}) An example of the attention matrices for all heads in a single layer (layer 8), given the above sentence as input. Phrases in the sentence are drawn as blocks in the matrix. (\textbf{b}) The change in attention, measured by relative entropy, between the unperturbed and perturbed attention weights, averaged over all out-of-phrase swaps. Columns are sorted independently by their value. (\textbf{c}) The head/layer-wise rank correlations ($\pm95\%$ confidence intervals) between distortion and tree distance after controlling for changes in attention, plotted against the uncontrolled rank correlations. Being below the diagonal indicates that the relationship between distortion and tree distance is partially explained by $\Delta a$.}
    \label{fig:fig7}
    \vspace{-5mm}
\end{figure}

First, we observe that the changes in the attention depend on the layer hierarchy when adjacent word swaps break the phrase boundary. Like the distortion, attention changes little or not at all in the early layers, and progressively more in the final layers (Fig.~\ref{fig:fig7}b). Furthermore, these changes are also positively correlated with syntactic distance in most cases, which suggests that representation's sensitivities to syntactic tree distance may primarily be due to changes in attention.

To see whether the changes in attention can in fact explain representational sensitivity to syntactic distance, we turned to the same partial linear regression model as before (\ref{SM:pmi}) to compute the the correlation between the representation's distortion and the tree distance between the swapped adjacent words, after controlling for changes in attention  ($\Delta a$). The correlations substantially reduced in the controlled case (Fig.~\ref{fig:fig7}c), which suggests that attention weights contribute to the representational sensitivity to syntactic tree distance; but the correlations are not eliminated, which suggests that distortions from the previous layer also contribute.

\section{Discussion and Conclusion}
\label{discussion}
In this paper, we used the representational change in response to perturbed input in order to probe the encoding of hierarchical phrasal structure in deep language models. We also use the same methods to show that changes in attention provide a partial explanation for perturbation-induced distortions. We find that the characteristic scale of word sequences, as measured by representational sensitivity to n-gram shuffling, grow along the model layers, similar to the increasing size of receptive fields found in sensory systems.

In language, a phrase within a sentence can serve as a conceptual unit, much like an object in a visual scene, thus motivating our perturbation-based probe for object-like representational sensitivity of phrases. We showed that BERT and its variants are indeed sensitive to the phrasal unit, as demonstrated by greater invariance to perturbations preserving phrasal boundaries compared to control perturbations which break the phrasal boundaries (Fig.~\ref{fig:fig2}-\ref{fig:fig7}). We also find that while the representational sensitivity to phrase boundaries grows across layers, this increase in sensitivity is more prominent when the phrase is broken by two adjacent words whose syntactic distance is far (i.e., when the broken phrase is more complex).

While our measure, inspired by prior work in neuroscience, measures the task-agnostic change in the neural population activity, we note that it's also possible to measure the amount of task-specific information in the representation by measuring the change in the performance or the cost function of supervised probes trained on top of the representation. To account for this, we compare our representational sensitivity measure with the changes in the performance of linguistic probes (Fig. ~\ref{fig:fig6}), and show that they are positively correlated. The probe sensitivity measure also bears a suggestive resemblance to the saliency map analysis \cite{simonyan2014deep} in machine learning, which is used to highlight the most output-sensitive regions within the input. In a similar spirit, one way of characterizing our results is that phrasal boundaries are regions of high saliency in hidden representations and, in deep layers, complex phrase boundaries are more salient than simple phrase boundaries. Exploring further the use of supervised probes and our input perturbations as a tool for layerwise probing of syntactic saliency is a promising direction for future work. Finally, \citet{sinha2021masked} recently found that masked language models pretrained on sentences that break natural word order (e.g. via n-gram shuffling) still perform quite well, even on supervised probes that probe for syntactic phenomena. It would be interesting to apply our perturbative analyses on such models to see if these models exhibit less sensitivity to the experimental vs. control setups (e.g. n-gram vs. phrase swaps), which may indicate that such models do not capture representational correlates of phrase structure in their representations despite their good performance on supervised probing tasks.

Our method and results suggest many interesting future directions. We hope that this work will motivate: (1) a formal theory of efficient hierarchical data representations in distributed features; (2) a search for the causal connection between attention structure, the representational geometry, and the model performance; (3) potential applications in network pruning studies; (4) an extension of the current work as a hypothesis generator in neuroscience to understand how neural populations implement tasks with an underlying compositional structure. 

\bibliographystyle{acl_natbib}
\bibliography{sample-base}



\appendix
\section{Appendix}
Here we go into further detail on our methods and data to aid in reproducibility.
\subsection{Additional details on the dataset}
In this section, we describe additional details of the manipulations done on the datasets.
\label{SM:dataset}
\paragraph{$n$-gram shuffling} 
For a given a sentence, we split it into sequential non-overlapping $n$-gram’s from left to right; if the length of the sentence is not a multiple of $n$, the remaining words form an additional $m$-gram, $m<n$. The list of the $n$-gram’s is randomly shuffled. Note that the 1-gram case is equivalent to a random shuffling of the words. 
In our analysis, we consider $n$-grams, with $n$ varying from 1 (i.e., individual words) to 7 and all the sentences have at least 10 words.

We provide here an example of $n$-gram shuffling.

\begin{itemize}
\item Original: The market 's pessimism reflects the gloomy outlook in Detroit
\item 1-gram : \textcolor{blue}{market} \textcolor{red}{pessimism} \textcolor{green}{the} \textcolor{orange}{'s} \textcolor{purple}{Detroit} \textcolor{pink}{in} \textcolor{black}{The} \textcolor{gray}{gloomy} \textcolor{magenta}{reflects} \textcolor{cyan}{outlook}
\item 2-gram : \textcolor{blue}{'s pessimism} \textcolor{red}{in Detroit} \textcolor{green}{The market} \textcolor{orange}{reflects the} \textcolor{purple}{gloomy outlook}
\item 3-gram : \textcolor{blue}{The market 's} \textcolor{red}{gloomy outlook in} \textcolor{green}{pessimism reflects the} \textcolor{orange}{Detroit}
\item 4-gram : \textcolor{blue}{in Detroit} \textcolor{red}{The market 's pessimism} \textcolor{green}{reflects the gloomy outlook}
\item 5-gram : \textcolor{blue}{the gloomy outlook in Detroit} \textcolor{red}{The market 's pessimism reflects}
\item 6-gram : \textcolor{blue}{outlook in Detroit} \textcolor{red}{The market 's pessimism reflects the gloomy}
\item 7-gram : \textcolor{blue}{in Detroit} \textcolor{red}{The market 's pessimism reflects the gloomy outlook }
\end{itemize}
\paragraph{Phrase swaps}
Using constituency trees from the Penn Treebank\cite{ptb}, we define phrases as constituents which don't contain any others within them. (See Fig.~\ref{fig:fig2}c or Fig.~\ref{fig:fig3}a in the main text.) Phrase swaps thus consist of swapping one phrase with another, and leaving other words intact.

To provide an appropriate control perturbation, we swap two disjoint $n$-grams, which are the same size as true phrases but cross phrase boundaries. 

\paragraph{Adjacent word swaps}
To better isolate the effect of broken phrase boundaries, we used adjacent word swaps. Adjacent words were chosen randomly, and one swap was performed per sentence.





\subsection{Partial linear regression}
\label{SM:pmi}
In order to control for uninteresting explanations of our results, we often make use of a simple method for regressing out confounds. Generally, we want to assess the linear relationship between $X$ and $Y$, when accounting for the (potentially non-linear) effect of another variable $Z$. In our experiments, $X$ is always the swap-induced distortion and $Y$ is the swap type, like integer-valued tree distance or binary-valued in/out phrase. We wish to allow $\mathbb{E}[Y \vert Z]$ and $\mathbb{E}[X \vert Z]$ to be any smooth function of $Z$, which is achieved by the least-squares solution to the following partially linear model:
\[ Y \sim \beta_x X + \mathbf{\beta_z} \cdot \mathbf{f}(Z) \Longleftrightarrow (Y - \mathbb{E}[Y \vert Z]) \sim \beta_x (X - \mathbb{E}[X \vert Z]) \]
where $\mathbf{f}(z)$ is a vector of several (we use 10) basis functions (we used cubic splines with knots at 10 quantiles) of $Z$. Both regressions have the same optimal $\beta_x$, but the one on the left is computationally simpler~\citep{econometrics}. The standard confidence intervals on $\beta_x$ apply.

Intuitively, the $\beta_x$ obtained by the partially linear regression above is related to the conditional correlation of $X$ and $Y$ given $Z$: $\rho (X, Y \vert Z)$. Like an unconditonal correlation, it will be zero if $X$ and $Y$ are conditionally independent given $Z$, but not necessarily \textit{vice versa} (both $X$ and $Y$ must be Gaussian for the other direction to be true). To compute conditional rank correlations (which assess a monotonic relationship between $X$ and $Y$), we rank-transform $X$ and $Y$ (this changes the confidence interval calculations).

We apply this method to swap size in Fig.~\ref{fig:fig2} and attentions in Fig.~\ref{fig:fig7}. In these supplemental materials, we will also report the results when $X$ is the binary in/out phrase variable, and $Z$ is PMI. The full $p$-values and coefficients of the uncontrolled and controlled regressions can be found in Table~\ref{SM:tab1}, where we observe that past layer 2, the $p$-value on phrase boundary is very significant ($p<10^{-12}$).


\begin{table*}\centering
\begin{tabular}{@{}rccccc@{}}\toprule
& \multicolumn{2}{c}{Without PMI} & \phantom{abc} & \multicolumn{2}{c}{With PMI}\\
\cmidrule{2-3} \cmidrule{5-6}
Layer & Coeff.$ \times 10^{-2}$ & p-value && Coeff. $\times 10^{-2}$ & p-value\\ \midrule
Emb. & $-0.21$ & $5.6 \times 10^{-5}$  && $-0.11$  & $9.4 \times 10^{-2}$\\
1 & $-0.11$ & $3.4 \times 10^{-2}$  && $-0.05$ & $4.2 \times 10^{-1}$\\
2 & $-0.74$ & $<10^{-16}$  && $-0.53$ & $2.12 \times 10^{-8}$\\
3 & $-1.6$ & $<10^{-16}$  && $-1.3$ & $2.2 \times 10^{-16}$\\
4 & $-2.0$ & $<10^{-16}$  && $-1.4$ & $4.4 \times 10^{-16}$\\
5 & $-2.1$ & $<10^{-16}$  && $-1.5$ & $8.8 \times 10^{-16}$\\
6 & $-2.4$ & $<10^{-16}$  && $-1.7$ & $<10^{-16}$\\
7 & $-2.6$ & $<10^{-16}$  && $-1.7$ & $1.6 \times 10^{-15}$\\
8 & $-3.4$ & $<10^{-16}$  && $-2.3$ & $<10^{-16}$\\
9 & $-3.8$ & $<10^{-16}$  && $-2.7$ & $<10^{-16}$\\
10 & $-4.1$ & $<10^{-16}$  && $-3.0$ & $<10^{-16}$\\
11 & $-3.8$ & $<10^{-16}$  && $-2.8$ & $<10^{-16}$\\
12 & $-4.2$ & $<10^{-16}$  && $-3.1$ & $<10^{-16}$\\
\bottomrule
\end{tabular}
\caption{Coefficients and p-values of the regular (`without PMI') and controlled (`with PMI') regressions of distortion against phrase boundary. }
\label{SM:tab1}
\end{table*}

\subsection{Supervised probes}
\label{SM:probes}
In this section, we describe the experiments based on the three linguistic tasks: parts of Speech (POS); grandparent tags (GP); and constituency tree distance. 

The POS and GP classifiers were multinomial logistic regressions trained to classify each word's POS tag (e.g. `NNP', `VB') and the tag of its grandparent in the constituency tree, respectively. If a word has no grandparent, its label is the root token `S'. The probes were optimized with standard stochastic gradient descent, 50 sentences from the PTB per mini-batch. 10 epochs, at $10^{-3}$ learning rate, were sufficient to reach convergence.

The distance probe is a linear map $\mathbf{B}$ applied to each word-vector $\mathbf{w}$ in the sentence, and trained such that, for all word pairs $i,j$, $\mathrm{TreeDist}(i,j)$ matches $\Vert \mathbf{B}(\mathbf{w}_i - \mathbf{w}_j) \Vert_2^2$ as closely as possible. Unlike the classifiers, there is freedom in the output dimension of $\mathbf{B}$; we used 100, although performance and results are empirically the same for any choice greater than $\sim 64$. Our probes are different from \cite{stanford_nlp} in two ways: (1) we use constituency trees, instead of dependency trees, and (2) instead of an L1 loss function, we use the Poisson (negative) log-likelihood as the loss function. That is, if $\lambda_{i,j} = \Vert \mathbf{B}(\mathbf{w}_i - \mathbf{w}_j) \Vert_2^2$, and $y_{i,j} = \mathrm{TreeDist}(i,j)$
\[ -l_{i,j} = y_{i,j}\log \lambda_{i,j} - \lambda_{i,j} - \log y_{i,j}! \]
Otherwise, the probes are trained exactly as in \cite{stanford_nlp}. Specifically, we used standard SGD with 20 sentences from the PTB in each mini-batch, for 40 epochs.

\paragraph{Evaluation}
A linear model is fit to maximize $p(y \vert \theta(\textbf{x}))$, with $p$ a probability function (multinomial for classifiers, Poisson for distance), and $\textbf{x}$ coming from the unperturbed transformer representation. We evaluate the model on $\tilde{\textbf{x}}$, which are the representations of the data when generated from a perturbed sentence. We take the average of $\log p(y \vert \theta(\textbf{x}_i)) - \log p(y \vert \theta(\tilde{\textbf{x}}_i))$ over all the data $i$ in all sentences. For example, all words for the classifiers, and all pairs of words for the distance probe. Concretely, we are just measuring the difference in validation loss of the same probe on the $\textbf{x}$ data and the $\tilde{\textbf{x}}$ data. But because the loss is an appropriate probability function, we can interpret the same quantity as a difference in log-likelihood between the distribution conditioned on the regular representation and that conditioned on the perturbed representation. Distortion is similarly computed using the full sentence, providing a number for each swap in each sentence.

\end{document}